\pdfoutput=1

\documentclass[11pt]{article}

\usepackage[final]{acl}

\usepackage{times}
\usepackage{latexsym}

\usepackage[T1]{fontenc}

\usepackage[utf8]{inputenc}

\usepackage{microtype}

\usepackage{inconsolata}

\usepackage{graphicx}

\usepackage{multirow}
\usepackage{enumitem}

\usepackage{comment}
\usepackage{bm}
\usepackage{makecell}
\usepackage{threeparttable}
\usepackage{xcolor}

\usepackage[normalem]{ulem}

\newcommand{\promptbox}[2]{%
\begin{center}
\fbox{%
\begin{minipage}{0.95\linewidth}
\small
\textbf{#1}

#2
\end{minipage}
}
\end{center}
}

\title{Detecting Sensitive Personal Information in Japanese \\
Pre-Training Corpora for Large Language Models}

\author{
  \textbf{Rei Minamoto\textsuperscript{1,2}},
  \textbf{Yusuke Oda\textsuperscript{2}},
  \textbf{Daisuke Kawahara\textsuperscript{1,2}}
\\
  \textsuperscript{1}Waseda University\\
  \textsuperscript{2}Research and Development Center for LLMs, National Institute of Informatics
\\
  \texttt{\{ray@akane., dkw@\}waseda.jp odashi@nii.ac.jp}
}

\begin{document}
\maketitle
\begin{abstract}
Sensitive personal information can appear in large-scale pre-training corpora for large language models (LLMs). Detecting and filtering such information is therefore essential to ensure compliance with privacy regulations and prevent unintended information leakage. However, in contrast to English and other languages, research into sensitive personal information has been limited in the Japanese language. In this study, we focus on sensitive personal data defined as special care-required personal information (SCPI) under Japan's Act on the Protection of Personal Information (APPI). We construct an SCPI dataset using LLM-based annotation and train machine learning models to rapidly detect SCPI in text. As a result, our SCPI classifier can effectively identify information related to SCPI. This study is the first to explore SCPI detection in Japanese text corpora, highlighting the challenges of accurate detection.
\end{abstract}

\section{Introduction}

Large-scale pre-training corpora are essential for building large language models (LLMs), and web crawling is a common way to construct them. Crawlers comprehensively gather publicly available data, although they may unintentionally collect personal information. If such content remains unfiltered within training corpora, there is a risk that LLMs may memorize it and subsequently leak it through their outputs \citep{huang-etal-2022-large}. 

Privacy laws in various countries impose stricter rules on sensitive attributes than on ordinary personal information. 
For example, the General Data Protection Regulation (GDPR) \citep{GDPR} defines them as special categories of personal data in Article 9, and 
the California Privacy Rights Act (CPRA), which amends the California Consumer Privacy Act (CCPA), defines sensitive personal information in California Civil Code Section 1798.140(ae) \citep{CCPA}. 

In Japan, the Act on the Protection of Personal Information (APPI) \citep{APPI} defines these attributes as special care-required personal information (SCPI) in Article 2. When SCPI remains in training corpora or is memorized by LLMs, the social and legal impacts of leakage will be even more severe. Therefore, corpora collected by crawling must be filtered for SCPI, and accurate technologies that detect such information are essential.

As described later in Section \ref{SCPI}, SCPI data required for training the classifier is subject to legal restrictions on its use, and therefore, no publicly available SCPI datasets exist. Consequently, it is necessary to independently construct a dataset while ensuring compliance with legal regulations. Corpora collected via web crawling are very large, and in some cases, corpora used for LLM pre-training can contain trillions of words. For this reason, an SCPI classifier needs to operate both effectively and at high speed.

\begin{figure}[t]
  \centering
  \includegraphics[width=\columnwidth]{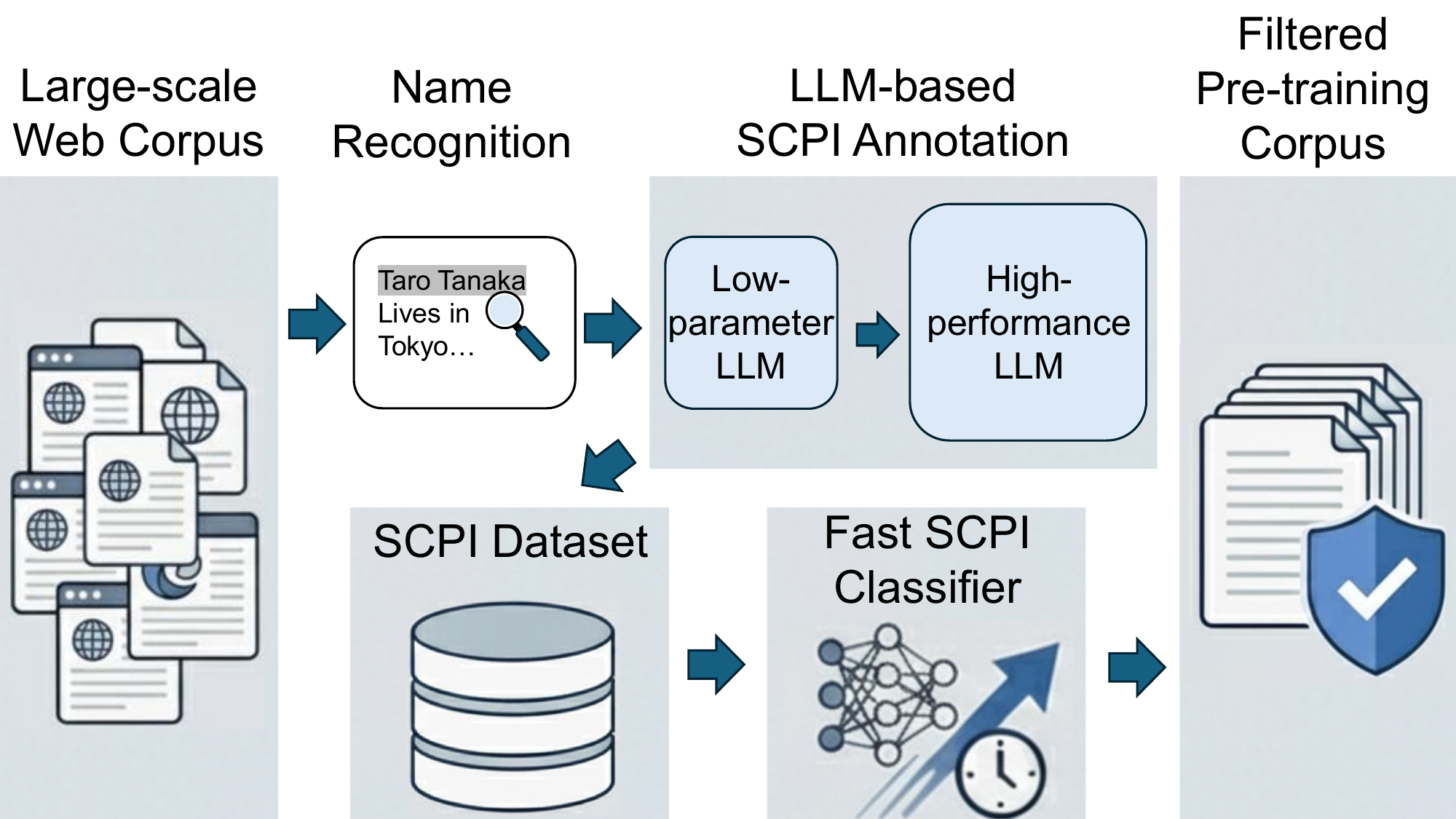}
  \caption{SCPI detection pipeline}
  \label{abstract}
\end{figure}

As shown in Figure \ref{abstract}, in this study, we apply LLM-based annotations to a text corpus in two stages to build an SCPI dataset. We then train various machine learning models on the annotated dataset to develop a fast SCPI classifier and evaluate their performance.

\section{Legal Background}
\subsection{Personal Information under APPI}
\label{Personal Information}
APPI defines personal information as ``information containing a name, date of birth, or other identifier or the equivalent which can be used to identify a specific individual'' or ``those containing an individual identification code.'' This latter category of individual identification code includes biometric information, passport numbers, and other similar identifiers. Non-text information contained in video or audio is not covered in this study.

Information that has already been made public in newspapers, on the Internet, and similar sources is also subject to protection as personal information. On the other hand, information about deceased people is not included in personal information.

\subsection{Special Care-required Personal Information under APPI}
\label{SCPI}
APPI defines SCPI as follows:
\begin{quote}
personal information as to an identifiable person's race, creed, social status, medical history, criminal record, the fact of having suffered damage by a crime, or other identifiers or their equivalent prescribed by Cabinet Order as those of requiring special care so as not to cause unjust discrimination, prejudice or other disadvantages to that person.
\end{quote}
Specifically, the following 11 items are defined as categories of SCPI in APPI and Cabinet Order under APPI \citep{CabinetOrder}: 
1. race, 2. creed, 3. social status, 4. medical history, 5. criminal record, 6. crime victimization, 7. disability, 8. diagnostic information, 9. medical treatment, 10. criminal proceedings, and 11. juvenile protection.

As mentioned above, SCPI is a type of personal information that requires special protection. Therefore, information constitutes SCPI only if it is personal information and includes information falling under the above categories.

With some exceptions, such as academic research purposes by academic institutions, the acquisition of SCPI and its provision to third parties generally requires the consent of the individual. However, this study is conducted for academic purposes by academic institutions and falls under the exception provisions based on APPI. Therefore, we annotate SCPI using LLMs and extract SCPI from the corpus while processing them appropriately and in compliance with the law. In carrying out this research, the measures described in the \nameref{Ethics} are a core part of the study and should be explicitly addressed.

Furthermore, on June 2, 2023, the Personal Information Protection Commission issued a warning regarding the use of generative AI services and specifically addressed OpenAI \citep{Warnings}. It emphasized the need to ensure that SCPI is not included in training datasets for machine learning, and if such data is included, it must be deleted or appropriately processed.

In this paper, we use the terms personally identifiable information (PII) and sensitive personal information (SPI) as general terms in related work. In the Japanese legal context, PII broadly corresponds to personal information under APPI, and SPI broadly corresponds to SCPI. However, SCPI is the specific legal category defined by APPI and the related Cabinet Order, and this paper focuses on SCPI.

\section{Related Work}
\label{Related}

As mentioned in the previous section, SCPI refers to personal information that includes sensitive information requiring special care. While a number of studies have addressed the detection or removal of PII from large-scale pre-training corpora, comparatively limited attention has been paid to SPI at the corpus level.

\paragraph{Detection of PII in Pre-Training Corpora}
Research on detecting PII in text corpora has received increasing attention. \citet{subramani-etal-2023-detecting} analyzed PII contained in the training corpora and compared detection methods.

Beyond analysis, in some recent pre-training corpora, PII filtering is applied during dataset construction. Common Corpus \citep{langlais2025commoncorpuslargestcollection} detected PII using Microsoft Presidio\footnote{\url{https://microsoft.github.io/presidio/}}, augmented with custom regular-expression patterns, and replaced detected PII. In contrast, Dolma \citep{soldaini-etal-2024-dolma} prioritized efficiency due to its scale and used carefully crafted regular expressions to detect and handle PII. However, these efforts explicitly target PII, while SPI is not explicitly discussed.

\paragraph{Detection of SPI}
\citet{fi14080228} developed a model to detect GDPR-defined special category data, trained on data from Reddit.
\citet{10031607} constructed a dataset containing GDPR-defined special category data through the annotation of information from a web corpus.
While several studies on SPI detection are grounded in legal definitions, there also exist studies that focus on sensitive information in LLM outputs without relying on such legal frameworks. \citet{zhang2025benchmark} proposed a benchmark dataset to evaluate semantically sensitive information in LLM outputs, which is related in scope but not directly comparable to our corpus-level SPI detection task.

Recent pre-training corpora such as Dolma3 adopt LLM-based filtering pipelines that target PII and may also remove sensitive attributes when they are linked to PII \citep{olmo2025olmo3}. However, this LLM-based filtering is applied only to specific subsets of the data, such as PDF documents, rather than to the entire pre-training corpus.

Despite this, research on SPI detection that is grounded in legal definitions and explicitly considers scalable, corpus-wide application to large-scale pre-training corpora remains limited.

\paragraph{Current Research on SCPI Detection}
In the Japanese context, there are no prior peer-reviewed studies that explicitly address SCPI detection. Moreover, systematic investigations of SCPI detection for large-scale Japanese text corpora have not been reported.

While there exist practical corpus curation pipelines\footnote{\url{https://github.com/matsuolab/jp-llm-corpus-pii-filter}} for Japanese LLM training data, these pipelines combine rule-based filtering based on surface-level patterns, such as Japanese personal names and explicit sensitive terms, with a Naive Bayes classifier for practical risk mitigation rather than exhaustive or semantically complete SCPI detection. Consequently, they are not directly comparable to methods designed for systematic SCPI detection evaluation.

Accordingly, this study is the first to systematically explore SCPI detection, addressing this critical gap in the literature.

\section{SCPI Dataset}
In this section, we describe how to construct a dataset for the SCPI classifier. The dataset described in this section consists of a collection of texts to be classified and a single SCPI label assigned to the text.

\subsection{SCPI Labels}
As mentioned in Section \ref{SCPI}, the APPI defines 11 categories of SCPI. Descriptions of each label are provided in Appendix \ref{SCPIlabel}. Additionally, we include non-SCPI, resulting in a total of 12 categories as labels.

\subsection{Dataset Construction Method}

\begin{figure}[t]
  \centering
  \includegraphics[width=0.9\columnwidth]{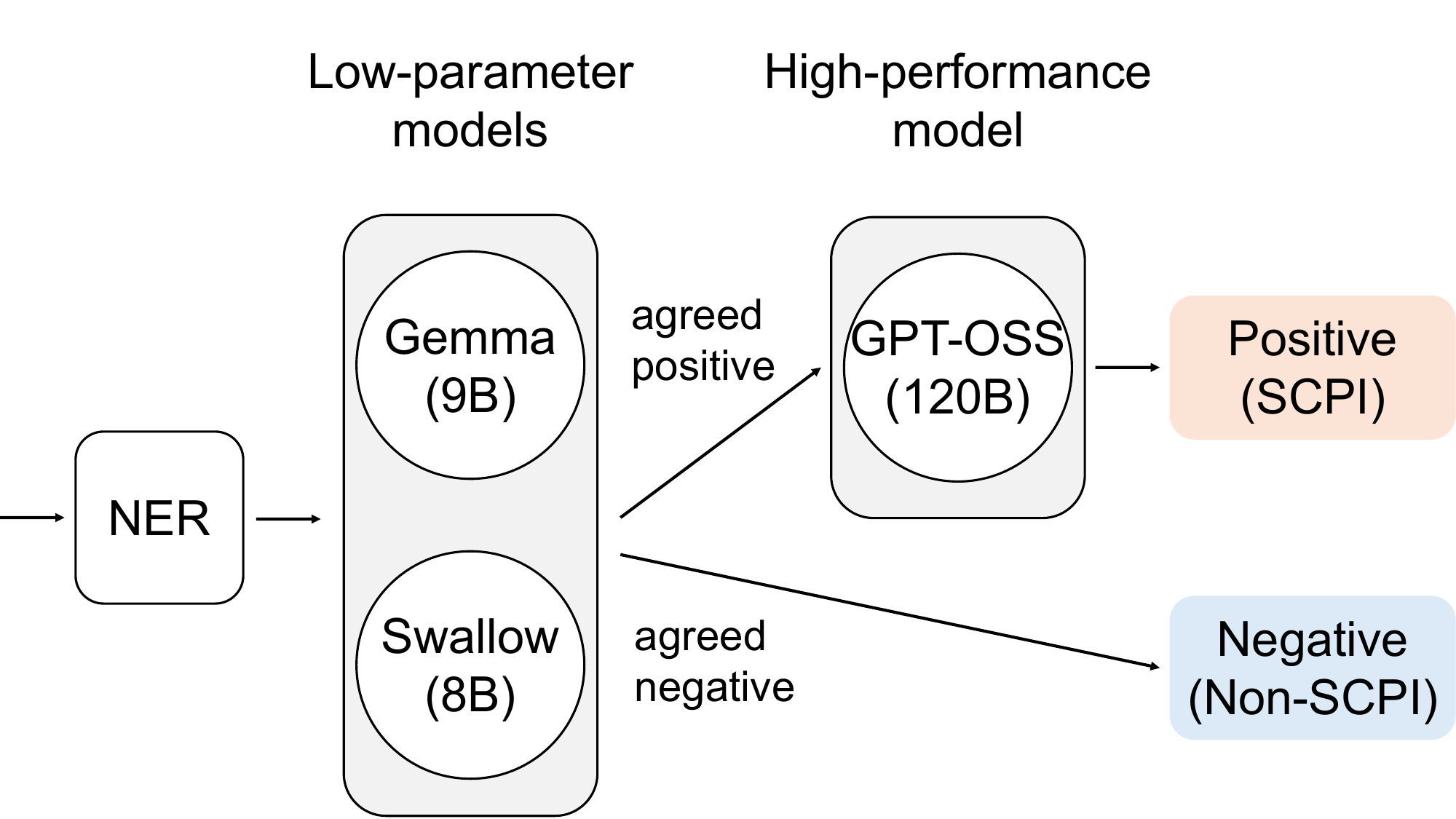}
  \caption{SCPI dataset construction pipeline}
  \label{pipeline}
\end{figure}

The dataset is constructed in two stages: (1) person name recognition and (2) SCPI annotation.

First, we perform person name recognition. Prior work has shown that person names are among the most frequently occurring forms of PII in web-derived corpora. For example, Presidio-based analysis reported that 97.3\% of detected PII in C4 and 96.1\% in the Pile correspond to names \citep{subramani-etal-2023-detecting}, making person names a practical entry point for candidate extraction. Under APPI, SCPI refers to personal information about identifiable individuals that includes sensitive information requiring special care. In this setting, a personal name can identify a specific individual on its own, whereas addresses or phone numbers are typically evaluated in combination with other information. We therefore use person name recognition as a practical candidate-reduction step to identify candidate texts during dataset construction.

Next, LLM-based annotation of SCPI is performed. To balance scalability and annotation quality, we adopt a two-stage annotation design. We first use low-parameter models to efficiently narrow down candidate texts, and then apply a high-performance model only to the filtered subset for final verification of SCPI labels.

For annotation, a prompt is used to determine whether a text contains SCPI. This prompt consists of an instruction, the SCPI definition, and the text to be annotated. The model returns an SCPI label corresponding to the text. Additionally, if the text is identified as containing SCPI, a confidence score ranging from 0 to 1 is returned. Details of the prompt are described in Appendix \ref{Prompt}. If the model produces inappropriate output that includes information other than SCPI labels and confidence scores, we handle it differently depending on the model. For low-parameter models, such outputs are treated as non-SCPI. For a high-performance model, where inappropriate outputs are less frequent, we perform manual annotation to ensure label accuracy.

\subsection{Dataset Construction Experiment}
\label{Dataset Construction Experiment}
The corpus used to extract texts containing SCPI
consisted of 2.61 million texts sampled from llm-jp-corpus-v3/ja/ja\_cc/level0\footnote{\url{https://gitlab.llm-jp.nii.ac.jp/datasets/llm-jp-corpus-v3/-/tree/main/ja/ja_cc/level0}}, a Japanese web corpus collected from Common Crawl. Each text had an average length of 2,474 characters.

For person name recognition, we used the NER model\footnote{\url{https://huggingface.co/llm-book/bert-base-japanese-v3-ner-wikipedia-dataset}}, detecting names in 1.06 million texts. The performance of the NER model used for person name recognition is reported in Appendix \ref{NER model performance}.

\begin{table}[t]
    \centering
    \small
    \begin{tabular}{lrc}
    \hline
    \textbf{Model} & \textbf{Count} & \textbf{Precision}\\
    \hline
    Swallow & 9,220 & 0.32\\
    Gemma & 72,262 & 0.11\\
    Gemma (0.8) & 15,722 &  0.16\\
    Swallow + Gemma (0.8) & 2,373 &  0.55\\
    \hline
    \end{tabular}
    \caption{Annotation results using low-parameter models}
    \label{swallow+gemma}
\end{table}

Among the open-weight LLMs that support Japanese, we used Llama 3.1 Swallow 8B Instruct v0.2 (Swallow) \cite{Fujii:COLM2024,Okazaki:COLM2024,ma:arxiv2025} and Gemma 2 9B IT (Gemma) \cite{gemma_2024} for annotating texts in which names were detected.
The number of detected texts containing SCPI and their precision, calculated based on manual review of 100 detection results, are shown in Table \ref{swallow+gemma}. Note that Gemma (0.8) selects texts with a confidence score of 0.8 or higher and Swallow + Gemma (0.8) selects texts that both models judge to be SCPI. Precision is calculated based on the condition that texts containing any SCPI are treated as positive examples, while texts without SCPI are treated as negative examples. Swallow + Gemma (0.8) achieved the highest precision with 2,373 cases, which were adopted as the result of the low-parameter models.

We used gpt-oss-120b \cite{openai2025gptoss120bgptoss20bmodel} as the high-performance model and applied it to the 2,373 candidate documents mentioned above. After manual correction of inappropriate outputs, 1,216 texts were identified as containing SCPI. During manual correction of inappropriate outputs, we observed a recurring group of cases, including references to sexual orientation and gender identity, that was not well captured by the original label set. We therefore introduced an additional category, labeled LGBT, for analysis. The breakdown of the resulting labels is shown in Table \ref{label}.

\begin{table}[t]
    \centering
    \small
    \begin{tabular}{lr}
    \hline
    \textbf{Label} & \textbf{Count} \\
    \hline
    race & 89\\
    creed & 50\\
    social status & 6\\
    medical history & 377\\
    criminal record & 76\\
    crime victimization & 42\\
    disability & 316\\
    diagnostic information & 9\\
    medical treatment & 42\\
    criminal proceedings & 143\\
    juvenile protection & 1\\
    LGBT & 65\\
    \hline
    \end{tabular}
    \caption{Breakdown of the resulting labels}
    \label{label}
\end{table}

\subsection{Post-processing and Analysis of the Dataset}
As shown in Table \ref{label}, the only labels with over 100 instances collected were medical history, disability, and criminal proceedings. Criminal record, crime victimization, and criminal proceedings may overlap and are difficult to classify accurately, and thus, we grouped them together as crime, resulting in a total of 261 instances. Accordingly, the categories used for classifier training were medical history (377), crime (261), and disability (316), for a total of 954 texts. The remaining SCPI categories were retained but not used for classifier training in this study because the number of collected instances was too small for robust training and evaluation.

\begin{table}[t]
    \centering
    \small
    \begin{tabular}{lccc}
    \hline
    \textbf{Label} & \textbf{Precision} & \textbf{Agree.} & $\bm{\kappa}$\\ 
    \hline
    medical history & 0.845 & 0.905 & 0.648\\
    criminal record & 0.921 & 0.961 & 0.734\\ 
    crime victimization & 0.690 & 0.690 & 0.350\\
    criminal proceedings & 0.979 & 0.972 & 0.322\\
    disability & 0.953 & 0.962 & 0.586\\
    \hline
    Total & 0.900 & 0.929 & 0.614\\
    \hline
    \end{tabular}    
    \caption{Annotation quality evaluation (Agree.:
agreement rate; $\kappa$: Cohen’s kappa)}
    \label{SCPIdataset evaluation}
\end{table}

We conducted a manual evaluation of the annotation quality of these 954 texts, and the results are presented in Table \ref{SCPIdataset evaluation}. Two annotators independently evaluated the labels, and the average precision across annotators was computed. In addition, inter-annotator agreement (IAA) was measured using agreement rate and Cohen’s kappa coefficient.

Overall, the annotation quality was high, with an average precision of 0.900, an agreement rate of 0.929, and a Cohen’s $\kappa$ of 0.614 across all labels, indicating substantial agreement. For crime victimization, precision and IAA were relatively lower, as the category includes boundary cases, such as large-scale events (e.g., war-related incidents) and cases that are less likely to be officially recognized or recorded as criminal offenses, including abuse and sexual harassment. The low $\kappa$ value for criminal proceedings is mainly attributable to severe class imbalance, despite the high agreement rate.

\begin{table}[t]
    \centering
    \small
    \begin{tabular}{lr}
    \hline
    \textbf{Label} & \textbf{Count} \\
    \hline
    medical history & 377\\
    crime & 261\\
    disability & 316\\
    non-SCPI & 8,586\\
    \hline
    \end{tabular}
    \caption{Breakdown of labels in the training dataset}
    \label{dataset}
\end{table}

Based on the above results, we use 954 cases from medical history, crime, and disability as positive examples to train the classifier described in the next sections. Negative examples were texts identified by both Swallow and Gemma as non-SCPI, as shown in Figure \ref{pipeline}. The proportion of texts containing SCPI in the entire corpus is extremely low. If the training data reflects this proportion, the classifier is likely to become biased toward negative examples. Therefore, as presented in Table \ref{dataset}, the positive examples were adjusted to constitute 1/10 of the training dataset.

Diagnostic information and medical treatment are typically held by medical institutions, and are unlikely to appear in web-collected data. Moreover, due to the nature of the data, personal information regarding juvenile protection is generally not made public. As a result, such information is expected to be scarce on the web, and even if present, is likely categorized as crime.

\section{SCPI Classification}
\label{SCPIclassifier}
\subsection{Classification Methods}
The SCPI dataset described in the previous section is used to train an SCPI classifier. This classifier categorizes the input text into one of four categories: medical history, crime, disability, or non-SCPI. Given that this classifier is applied to large-scale corpora, not only the performance but also its speed needs to be considered. Because full-corpus inference with LLMs is computationally expensive, we use LLMs only for high-quality annotation and train lightweight models for efficient large-scale inference. We examined efficient traditional machine learning models and a BERT-based Transformer model \citep{NIPS2017_3f5ee243}.

\subsection{Experiments}
To evaluate SCPI classification models, we perform cross-validation and manual evaluation on a large-scale corpus. We also examine model stacking to assess potential complementary effects across models.

\subsubsection{Individual Model Evaluation}
\paragraph{Model Description}
The 9,540 instances in Table \ref{dataset} were used to train the classifier. We used two types of models: traditional machine learning models and a Transformer model. The traditional machine learning models were SVC from scikit-learn \cite{scikit-learn}, LGBMClassifier (LGBM) from lightgbm \cite{LightGBM}, and nn.Module (NN) from PyTorch \cite{Ansel_PyTorch_2_Faster_2024}. The Transformer model was modernbert-ja-310m (ModernBERT) \cite{modernbert-ja}, which is a BERT-based classifier. 

For text representation, we used vectorization techniques for all models except ModernBERT, which directly performs classification. Specifically, we used both lightweight approaches (Doc2Vec from gensim \cite{rehurek_lrec}, TfidfVectorizer from scikit-learn \cite{scikit-learn}) and a Transformer model (ruri-v3-310m \cite{Ruri}, which is a Japanese embedding model).
The resulting vectors are then fed into models for classification. The detailed settings for models and vector representations are provided in Section~\ref{modelsettings}.

\begin{figure*}[t]
  \centering
  \includegraphics[width=0.85\linewidth]{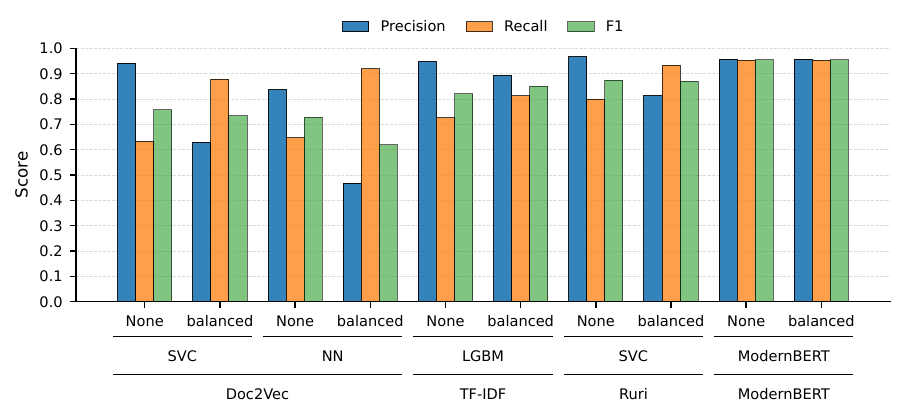}
  \caption{Cross-validation results for each base model on the training data}
  \label{result1}
\end{figure*}

\paragraph{Cross-Validation Performance Evaluation}
Figure \ref{result1} presents the results of 5-fold cross-validation on the training data. When class weight was set to None, no special adjustments were applied. Setting it to balanced automatically corrected class imbalance by adjusting weights based on the occurrence frequency of each class. Precision and recall are calculated based on the condition that texts containing any SCPI are treated as positive examples, while texts without SCPI are treated as negative examples. All evaluation metrics were calculated as the average values across all folds. In this experiment, the F1 score is used as the primary evaluation metric. However, since preventing SCPI from being overlooked is important, recall is prioritized over precision.

The best-performing model was ModernBERT, with all metrics exceeding 0.95. The second-best model was SVC with Ruri. When comparing traditional machine learning models, LGBM with TF-IDF had the best performance. Additionally, when the class weight was set to balanced, precision decreased and recall improved.

\begin{table*}[t]
    \centering
    \small
    \tabcolsep 5.5pt
    \begin{tabular}{cccccccccccc}
    \hline
    \multirow{2}{*}{\textbf{Vector}} & \multirow{2}{*}{\textbf{Model}} & \multirow{2}{*}{\textbf{Label}} & \multicolumn{3}{c}{\textbf{Annotator1}} & \multicolumn{3}{c}{\textbf{Annotator2}} & \multirow{2}{*}{\textbf{Agree.}} \\
    \ & & & \textbf{SCPI} & \textbf{Related} & \textbf{Unrelated} & \textbf{SCPI} & \textbf{Related} & \textbf{Unrelated} & \\
    \hline
    \multirow{4}{*}{Doc2Vec} & \multirow{4}{*}{\makecell{SVC\\(None)}} & medical history & 0.32 & 0.52 & 0.16 & 0.24 & 0.64 & 0.12 & 0.84 \\
    \ & & crime & 0.36 & 0.52 & 0.12 & 0.28 & 0.60 & 0.12 & 0.66 \\
    \ & & disability & 0.24 & 0.56 & 0.20 & 0.40 & 0.42 & 0.18 & 0.68 \\
    \cline{3-10}
    \ & & average & 0.31 & 0.53 & 0.16 & 0.31 & 0.55 & 0.14 & 0.73\\
    \hline
    \multirow{4}{*}{TF-IDF} & \multirow{4}{*}{\makecell{LGBM\\(balanced)}} & medical history & 0.16 & 0.76 & 0.08 & 0.14 & 0.78 & 0.08 & 0.86 \\
    \ & & crime & 0.36 & 0.56 & 0.08 & 0.36 & 0.52 & 0.12 & 0.74 \\
    \ & & disability & 0.16 & 0.62 & 0.22 & 0.16 & 0.78 & 0.06 & 0.76 \\
    \cline{3-10}
    \ & & average & 0.23 & 0.65 & 0.13 & 0.22 & 0.69 & 0.09 & 0.79\\
    \hline
    \multirow{4}{*}{-} & \multirow{4}{*}{\makecell{ModernBERT\\(None)}} & medical history & 0.38 & 0.54 & 0.08 & 0.32 & 0.64 & 0.04 & 0.86 \\
    \ & & crime & 0.56 & 0.36 & 0.08 & 0.52 & 0.42 & 0.06 & 0.82 \\
    \ & & disability & 0.48 & 0.50 & 0.02 & 0.52 & 0.48 & 0.00 & 0.86 \\
    \cline{3-10}
    \ & & average & 0.47 & 0.47 & 0.06 & 0.45 & 0.51 & 0.03 & 0.85\\
    \hline
    \multirow{4}{*}{-} & \multirow{4}{*}{gpt-oss-120b} & medical history & 0.62 & 0.24 & 0.14 & 0.56 & 0.30 & 0.14 & 0.76 \\
    \ & & crime & 0.66 & 0.26 & 0.08 & 0.68 & 0.24 & 0.08 & 0.84 \\
    \ & & disability & 0.84 & 0.08 & 0.08 & 0.82 & 0.14 & 0.04 & 0.86 \\
    \cline{3-10}
    \ & & average & 0.71 & 0.19 & 0.10 & 0.69 & 0.23 & 0.09 & 0.82\\
    \hline
    \end{tabular}   
    \caption{Results of SCPI extraction using classifiers (Agree.:
agreement rate)}
    \label{result2}
\end{table*}

\paragraph{Manual Performance Evaluation}
To evaluate the practical performance, we manually assess the detection results obtained by applying models to a large-scale corpus. Here, the large-scale corpus refers to the 1.06 million Japanese web texts from llm-jp-corpus-v3 in which person names were detected by the NER model, as described in Section \ref{Dataset Construction Experiment}. To prevent data leakage, texts included in the final SCPI dataset were excluded from this large-scale evaluation. The detected texts are manually categorized as follows: ``SCPI'' for texts containing SCPI, ``Related'' for texts that do not contain SCPI but include information related to medical history, crime, or disability (such as medical care or welfare for the disabled), and ``Unrelated'' for texts that do not contain SCPI or any related information. We evaluate better-performing traditional machine learning models (SVC and LGBM) and the Transformer model (ModernBERT), as identified through cross-validation. In addition, we evaluate gpt-oss-120b in a zero-shot setting as a baseline. Two annotators independently evaluated 50 texts for each label across all models.

Table \ref{result2} shows the results of manual evaluation. Detailed evaluation setups are reported in Appendix \ref{AppendixClassifier}, and examples of extracted SCPI are provided in Appendix \ref{SCPIexample}.
Because determining the absence of SCPI requires careful review of entire texts, agreement rate tends to be moderate.

Gpt-oss-120b detected the largest proportion of ``SCPI'' instances. ModernBERT detects ``SCPI'' more frequently than SVC and LGBM, while the proportion of ``SCPI'' is consistently lower across all models than the precision reported in Figure \ref{result1}.

\begin{figure}[t]
  \centering
  \includegraphics[width=\columnwidth]{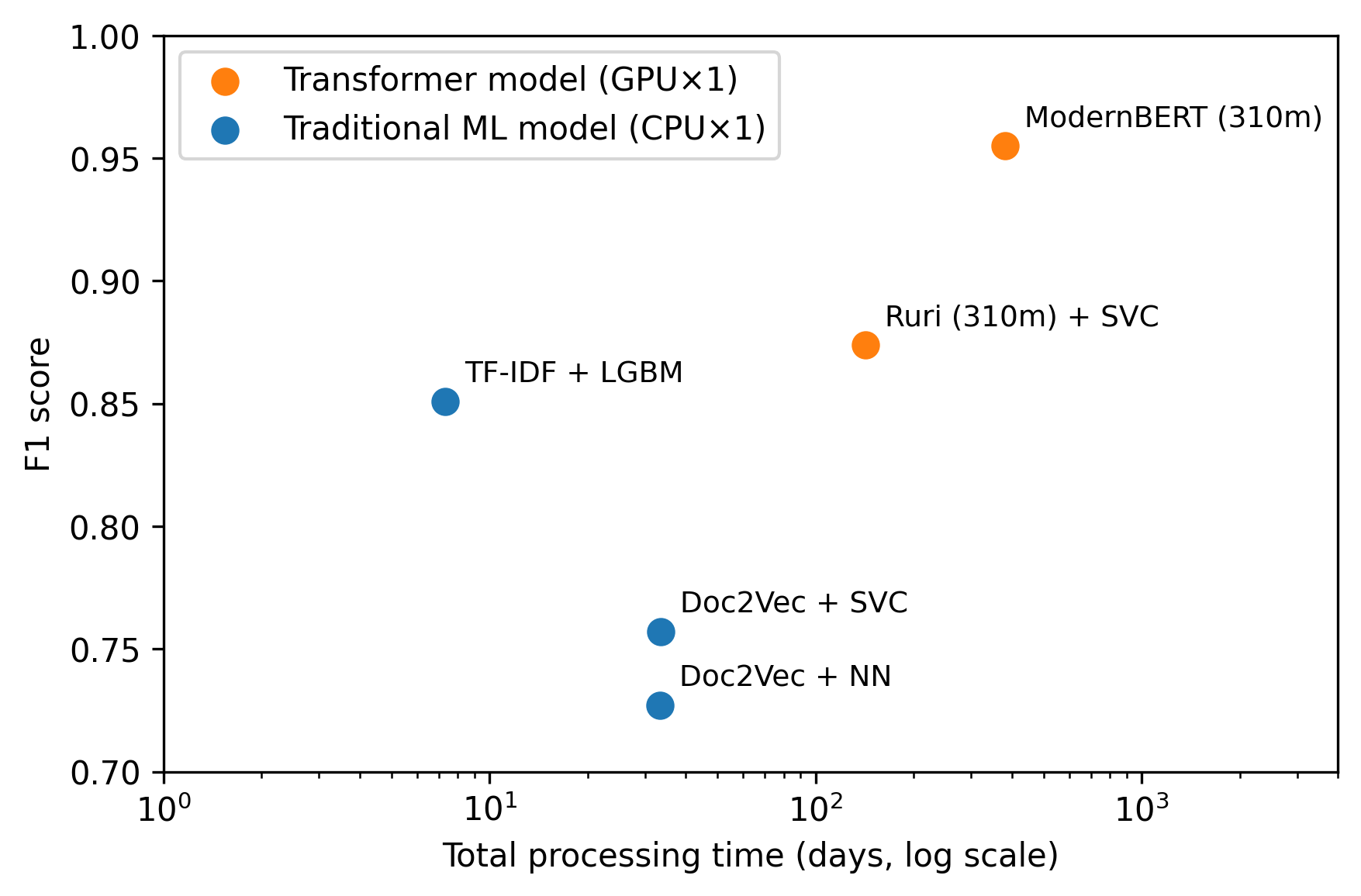}
  \caption{Processing speed for full-corpus inference}
  \label{speed}
\end{figure}

\paragraph{Processing Speed Comparison}
Finally, we compare the processing speed of each model. Based on the average processing speed of 10,000 texts, we estimate the total time required to process the entire corpus containing 184 million texts. The results are shown in Figure \ref{speed}. As the models were not optimized, the results should be viewed as approximate indicators of performance trends.

Doc2Vec and TF-IDF were run on a single CPU core, while Ruri and ModernBERT were executed using a single GPU. Even without parallelization, classical models are approximately 4 to 50 times faster than Transformer models. Considering the cost difference between CPUs and GPUs as well as the potential for parallelization, the practical speed advantage in real-world applications could reach several tens to hundreds of times.

For instance, LGBM with TF-IDF, when run on 8 CPU cores, can process the entire corpus within a day. In contrast, ModernBERT requires more than a month to process the same data, even with 8 GPUs. Because Transformer models impose significant computational and cost burdens, we adopt faster machine learning models in this study.

\begin{figure}[t]
  \centering
  \includegraphics[width=0.7\columnwidth]{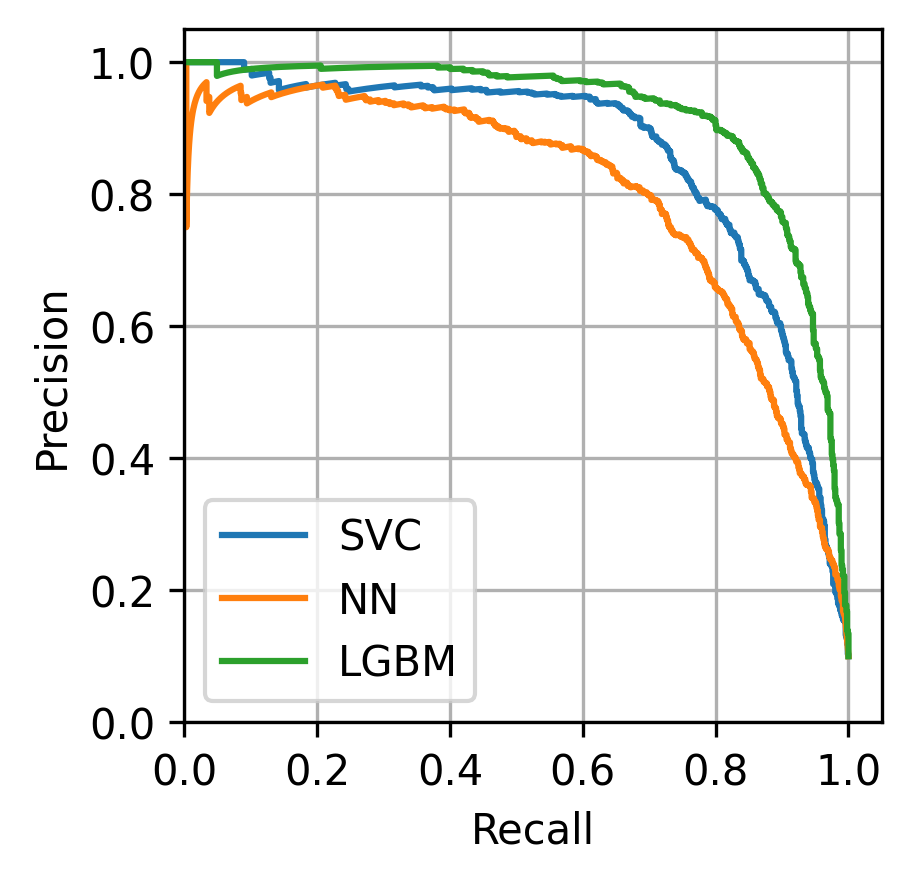}
  \caption{Precision-Recall curves for base models}
  \label{prec-rec}
\end{figure}

\subsubsection{Stacking Model Evaluation}
To perform stacking, the models with the highest F1 score were selected as the base models. Specifically, the following three machine learning models were selected:
\begin{itemize}
    \item SVC: class weight = None, 
    \item NN: class weight = None, and
    \item LGBM: class weight = balanced.
\end{itemize}

Figure \ref{prec-rec} shows Precision-Recall curves for base models. As shown in the figure, the precision is relatively high. Therefore, stacking is performed by taking the union of the identified results to improve recall. In other words, if any of the base models classifies the text as SCPI, it is considered SCPI. 

Table \ref{OR model} presents the performance of the base models and the stacked models. The combination of SVC and LGBM achieved the highest F1 score. To further improve performance, both models were tuned with Optuna \cite{optuna_2019}, with the F1 score set as the optimization objective. The F1 score was calculated based on the condition that three SCPI categories -- medical history, crime, and disability -- are treated as positive examples, and non-SCPI is treated as negative examples. LGBMClassifier cannot use this F1 score for the early stopping evaluation, which is essential for efficient training, so lgb.train was used instead.

\begin{table}[t]
    \centering
    \small
    \begin{tabular}{cccc}
    \hline
    \textbf{Model} & \textbf{Precision} & \textbf{Recall} & \textbf{F1}\\
    \hline
    SVC & 0.942 & 0.633 & 0.757\\
    NN & 0.837 & 0.647 & 0.727\\
    LGBM & 0.893 & 0.812 & 0.851\\
    SVC + NN & 0.847 & 0.724 & 0.780\\
    SVC + LGBM & 0.869 & 0.869 & 0.869\\
    LGBM + NN & 0.816 & 0.892 & 0.852\\
    SVC + LGBM + NN & 0.815 & 0.900 & 0.855\\
    \hline
    \end{tabular}
    \caption{Stacking results}
    \label{OR model}
\end{table}

\begin{table}[t]
    \centering
    \small
    \begin{tabular}{cccc}
    \hline
    \textbf{Model} & \textbf{Precision} & \textbf{Recall} & \textbf{F1}\\
    \hline
    SVC & 0.913 & 0.732 & 0.812\\
    LGBM & 0.883 & 0.856 & 0.869\\
    SVC + LGBM & 0.845 & 0.911 & 0.877\\
    \hline
    \end{tabular}
    \caption{Tuning results}
    \label{tunemodel}
\end{table}

Tuning results are presented in Table \ref{tunemodel}. The performance improved significantly compared to the default parameters. Although the F1 score did not increase substantially after stacking, recall, which is critical in this study, improved by 4.2\%. The hyperparameters are provided in Section \ref{hyparaopt}.

\begin{figure*}[t]
  \centering
  \begin{minipage}{0.45\linewidth}
    \centering
    \includegraphics[width=\linewidth]{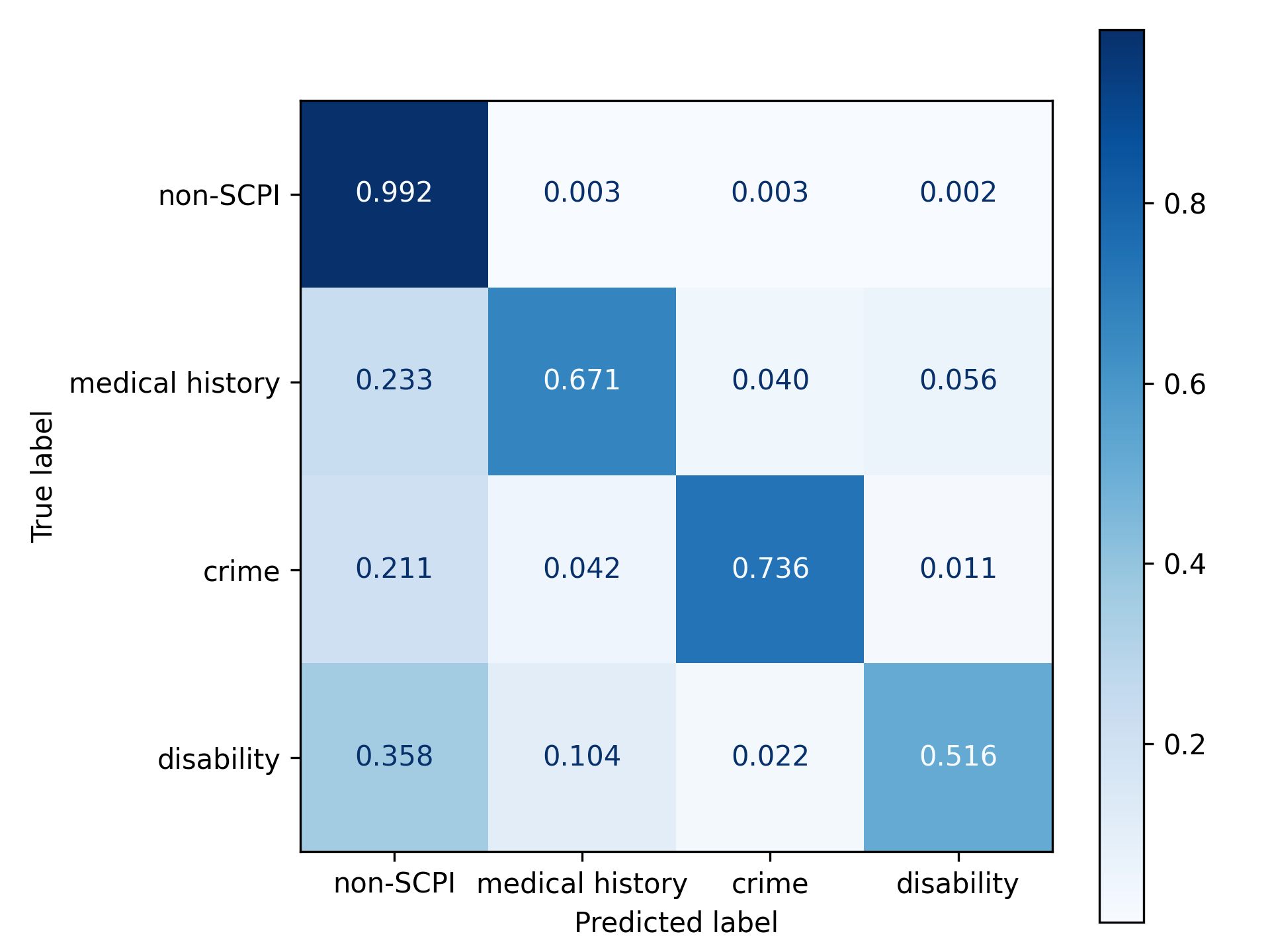}
    \caption{Normalized confusion matrix for SVC}
    \label{conf_tuneSVC}
  \end{minipage}\hfill
  \begin{minipage}{0.45\linewidth}
    \centering
    \includegraphics[width=\linewidth]{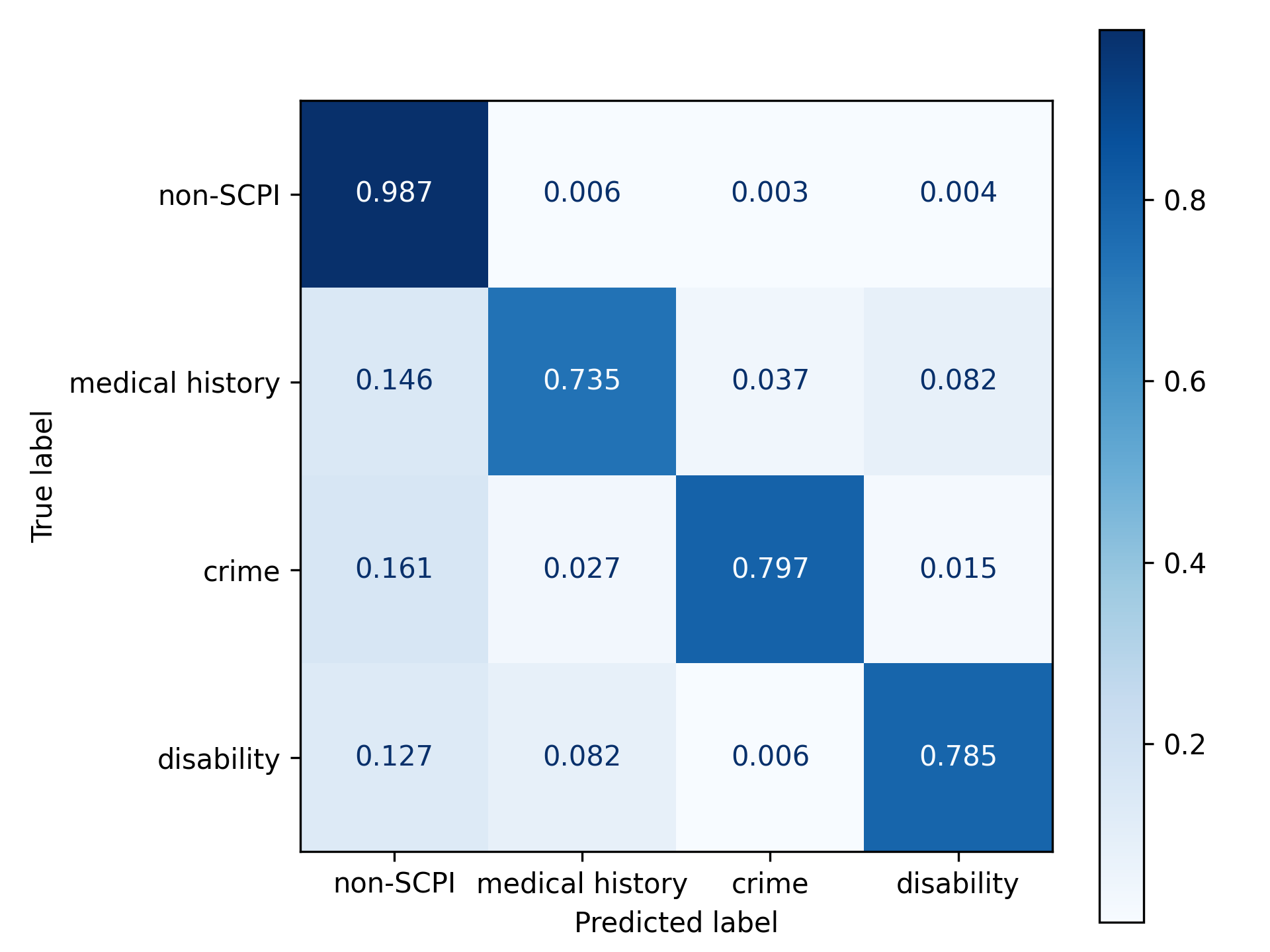}
    \caption{Normalized confusion matrix for LGBM}
    \label{conf_tuneLGBM}
  \end{minipage}
\end{figure*}

\begin{table*}
    \centering
    \small
    \begin{tabular}{ccc}
    \hline
    \textbf{Label} & \textbf{Related Words} & \textbf{Unrelated Words}\\
    \hline
    medical history & illness, treatment, hospital, cancer, symptom & present, later, time\\
    crime & suspicion, arrest, incident, trial, police & wish, introduction, place, year\\
    disability & disability, syndrome & person, sport, tournament, car, word\\
    \hline
    \end{tabular}
    \caption{Words with high absolute SHAP values}
    \label{SHAP}
\end{table*}

\subsection{Analysis}
\label{Results Analysis}

We analyze the classification performance and characteristic errors of each label. Figure \ref{conf_tuneSVC} and Figure \ref{conf_tuneLGBM} show the proportions of predicted labels and gold labels for the tuned models.

SVC tends to misclassify disability as non-SCPI more often than other labels. Among SCPI categories, disability is frequently mislabeled as medical history. In contrast, LGBM shows higher accuracy in classifying disability labels than SVC, with a higher likelihood of correctly identifying disability as SCPI. However, confusion between medical history and disability remains consistent.

Furthermore, we analyze the most influential words for each SCPI label in LGBM using SHAP values generated by TreeExplainer \cite{lundberg2020local2global}. Table \ref{SHAP} shows words with high absolute SHAP values translated from Japanese. Note that the table is not exact because some words cannot be translated. Although all the words in the table are within the top 15, it can be seen that many unrelated words are also included.

\subsection{Discussion}

The reason why the proportion of ``SCPI'' in Table \ref{result2} is lower than the precision in Figure \ref{result1} is likely because the current classifier can correctly identify SCPI-related category information but struggles to identify SCPI instances. Notably, the combined proportion of ``SCPI'' and ``Related'' matches the precision in Figure \ref{result1}, suggesting that the classifier performs well in detecting SCPI-related category information, consistent with Figure \ref{result1}. This gap may also be partly explained by the construction of negative examples. Because negative examples were selected based on agreement between two low-parameter models after NER-based filtering, this setting may introduce selection bias and lead to more optimistic cross-validation results.

Based on this observation, it is expected that more accurate judgments can be achieved by applying this classifier in the first stage to classify the SCPI-related category information, and then applying a second stage classifier to distinguish between ``SCPI'' and ``Related.'' Because the number of texts is reduced by rough filtering in the first stage, it is possible to use a higher-performance model in a practical amount of time in the second stage. Although Transformer models detect a higher proportion of ``SCPI'' instances, the combined proportion of ``SCPI'' and ``Related'' remains comparable across models. For this purpose, fast traditional machine learning models are suitable for the first stage filtering, enabling scalable extraction of SCPI-related category information.

In Section \ref{Results Analysis}, both models had the tendency to confuse the labels medical history and disability. This is thought to be due to the high correlation between these two categories. Although a single label is assigned to each text in this study, in reality, there are texts with multiple labels, suggesting that the model was able to correctly identify them. SVC also had the characteristic of low accuracy in identifying disabilities. Since many texts identified as disabilities contained information on caregiving and child-rearing, it is thought that the model was not limited to disabilities, but had learned texts related to a wider range of welfare. Conversely, LGBM with TF-IDF had high accuracy in identifying disabilities, which confirms that each vector learns different features and complements each other.

SHAP analysis revealed that many of the top-ranked features were semantically unrelated to the gold label, indicating room for further improvement. This issue arises because only a small portion of the training texts actually correspond to SCPI, while the majority are noise. To further improve performance, it is necessary to accurately extract the SCPI-relevant parts. However, SCPI requires both identifiable information and category information, which are often located far apart. In addition, it is necessary to determine whether the two elements are actually related, which requires a deep understanding of their semantic relationship. Although it is extremely difficult to accurately extract the relevant parts using LLMs, a high-performance model may be able to handle this task. Though time and cost are intensive, manual annotation is a reliable fallback for addressing the issue.

\section{Conclusion and Future Work}
In this study, we constructed an SCPI dataset and an SCPI classifier, and confirmed its performance. Traditional machine learning models demonstrated high accuracy in detecting SCPI-related category information, indicating its usefulness in the first stage of filtering, which classifies SCPI together with other related information. In addition, it operates at a lower cost than classification using LLMs and Transformer models, demonstrating advantages in terms of time and cost for processing large-scale corpora. 

To improve SCPI classification performance, we plan to extend the pipeline to handle PII other than names and use other labels that were not available in this study. We also plan to combine LLM-based annotation and manual annotation so that the two approaches complement each other, allowing us to capture SCPI that an LLM alone might miss and incorporate them back into the training data.

\section*{Limitations}
In this study, we mainly focus on name-based personal information when constructing the dataset, so this approach does not fully address SCPI in which individuals can be identified from information other than names, such as addresses and email addresses. Personal identification codes are also outside the scope of this research. These exclusions were motivated by the need to develop a filter for the most frequent pattern observed in Japanese web corpora containing a full name and sensitive categories. 

Although the corpus contains texts referring to SCPI categories such as race, creed, and social status, we did not evaluate the classifier’s performance on these categories in this study. In addition, during manual correction, we observed a recurring group of cases involving sexual orientation and gender identity. While these cases do not constitute SCPI categories under APPI, we treated them as an additional category, labeled LGBT, for analysis.

Since we also planned to outsource the manual annotation, we first prepared a pilot dataset to train classifiers that assist in collecting additional data for the categories not covered by the dataset described in this paper.

For the race, creed categories, as well as the additional LGBT category, we enlarged the source dataset and constructed an additional pilot dataset using parts of the original extraction pipeline combined with manual annotation. In contrast, even after expanding the source dataset, we still lacked sufficient texts for social status; therefore, we constructed a pilot dataset by extracting candidate sentences for this category using pattern matching and manual annotation.

The classifiers trained on these supplementary data were prepared exclusively for making an annotation dataset so they were optimized with precision as the primary objective. Because such a precision-oriented model falls outside the scope of this paper—which focuses on detecting SCPI—we do not discuss them further.

Additionally, since the dataset is constructed using LLM-based annotations, there is a limitation: any SCPI that cannot be detected by an LLM cannot be recognized as SCPI by the classifier.

\section*{Ethical Considerations}
\label{Ethics}

This research meets the ethical standards of the affiliated institution and has been conducted appropriately. In this study, we detect SCPI in the published corpus. The collected data will be used only by the research team and will not be made public. As described in Section \ref{SCPI}, this study is conducted for academic research purposes by an academic institution, and therefore falls under the exception provision of APPI Article 18, which allows the acquisition and processing of SCPI without obtaining the consent of the individual.

SCPI annotation is performed using LLMs. All models are run in a strictly local environment, where no data leaves our infrastructure.

The use policy for each LLM used in Section \ref{Dataset Construction Experiment} requires that the handling of SCPI be in compliance with applicable laws, and this research complies with these terms \cite{Llama,Gemma}.

While this research aims to promote privacy protection, we acknowledge that the trained SCPI classifier may also be misused to identify and extract SCPI from corpora for malicious purposes. Therefore, the decision to publicly release the trained model will be made with great caution, considering potential misuse, social impact, and legal compliance. Moreover, while there are plans to outsource manual annotation, they will comply with laws and regulations and handle the data with the utmost care. 

\bibliography{custom}

\appendix

\section{SCPI Definitions and Label Descriptions Used in Annotation}
\label{SCPIlabel}
The items of SCPI are defined in detail in the guidelines. The translated summary below corresponds to the SCPI definition and label descriptions used in the annotation prompts in Appendix B. For readability, we present the shared definitions here separately rather than repeating them in each prompt.

\subsection{SCPI Definition}
Special care-required personal information refers to personal information that requires special care so as not to cause unjust discrimination, prejudice or other disadvantages. It includes information related to the following items (1 to 11).
However, information that merely suggests the information in items 1 to 11 (e.g., information related to the purchase or borrowing of books on religion) is not considered special care-required personal information.

\subsection{Label Description}
\begin{description}
\item[race]: race, descent, and national or ethnic origin, excluding nationality or skin color
\item[creed]: an individual's fundamental beliefs, including both ideology and faith
\item[social status]: a position inherently tied to an individual's situations, which cannot be easily changed, excluding occupational status or educational background (e.g., being born out of wedlock or being from a Buraku community)
\item[medical history]: a history of having a particular illness (e.g. cancer, schizophrenia, etc.)
\item[criminal record]: a confirmed criminal conviction
\item[crime victimization]: the physical, mental and financial damage caused by a crime
\item[disability]: information on disabilities, including physical, intellectual, mental, developmental, or illness-related disabilities
\item[diagnostic information]: results of health checkups and other tests, including genetic tests, that reveal a person's health condition
\item[medical treatment]: information obtained from health guidance, medical treatment, or dispensing of medication, including the fact that a person received such guidance, treatment, or dispensing
\item[criminal proceedings]: the fact that a criminal proceeding has been conducted for a person as a suspect or defendant, excluding questioning or witness examination conducted for the investigation of another person
\item[juvenile protection]: the fact that a procedure was conducted for a juvenile delinquent in a juvenile protection case
\end{description}

\begin{table*}[t]
    \centering
    \small
    \tabcolsep 5.0pt
    \begin{tabular}{lcccccccc}
    \hline
    \multirow{2}{*}{\textbf{Label}} & \multicolumn{3}{c}{\textbf{Annotator1}} & \multicolumn{3}{c}{\textbf{Annotator2}} & \multirow{2}{*}{\textbf{Agree.}} & \multirow{2}{*}{$\bm{\kappa}$}\\
    \ & \textbf{SCPI} & \textbf{Name missing} & \textbf{Not sensitive} & \textbf{SCPI} & \textbf{Name missing} & \textbf{Not sensitive} & & \\
    \hline
    medical history & 0.862 & 0.103 & 0.034 & 0.828 & 0.135 & 0.037 & 0.905 & 0.648\\
    criminal record & 0.921 & 0.039 & 0.039 & 0.921 & 0.066 & 0.013 & 0.961 & 0.734\\ 
    crime victimization & 0.714 & 0.143 & 0.143 & 0.667 & 0.167 & 0.167 & 0.690 & 0.350\\
    criminal proceedings & 0.972 & 0.021 & 0.007 & 0.986 & 0.014 & 0.000 & 0.972 & 0.322\\
    disability & 0.965 & 0.019 & 0.016 & 0.940 & 0.022 & 0.038 & 0.962 & 0.586\\
    \hline
    Total & 0.911 & 0.060 & 0.029 & 0.889 & 0.075 & 0.036 & 0.929 & 0.614\\
    \hline
    \end{tabular}    
    \caption{Detailed annotation quality evaluation (Agree.: agreement rate; $\kappa$: Cohen's kappa)}
    \label{manual evaluation2}
\end{table*}

\section{Prompt Details}
\label{Prompt}
The annotation prompts consisted of task instructions, the SCPI definition and label descriptions summarized in Appendix A, output constraints, and the input text. This appendix describes the task-specific instruction design of each prompt.

\subsection{Prompt for Low-Parameter Models}
For low-parameter models, accuracy tended to decrease when annotation was performed in a single process. To address this issue, annotation  was performed in three stages below.

\promptbox{Prompt 1: SCPI detection}{
Analyze the input text and determine whether it contains SCPI.

\begin{itemize}
    \item If it does, output the corresponding category label (1--11) together with a confidence score (0--1).
    \item If it does not, output only 0.
\end{itemize}
}

\promptbox{Prompt 2: Name association}{
Analyze the input text and determine whether it contains a person name associated with SCPI.

\begin{itemize}
    \item If it does, output only 1.
    \item If it does not, output only 0.
    \item If the text does not contain SCPI, output only 0.
\end{itemize}
}

\promptbox{Prompt 3: Identifiable living individual check}{
Determine whether the following text contains a full name that identifies a living individual.

\begin{itemize}
    \item Output 1 if it does, and 0 otherwise.
\end{itemize}

\textbf{Notes:}
\begin{itemize}
    \item A full name refers only to a form in which both the family name and given name are present.
    \item Stage names and pen names are also treated as full names if they can identify a specific individual.
    \item A family name alone or a given name alone is not treated as a full name.
    \item Names of fictional characters and deceased individuals are not treated as full names.
    \item Romanized or katakana-rendered pseudonyms are not treated as full names.
\end{itemize}
}

\subsection{Prompt for gpt-oss-120b}
gpt-oss-120b maintained high accuracy even when annotation was performed in a single process. Therefore, the annotation was executed in a single run. Additionally, this model is capable of accurately handling multiple instructions. Leveraging this capability, instructions were added to the prompt to determine whether the individual associated with the SCPI is merely a related party, compared to the prompt presented in the previous section. Below, we present the instruction part of the prompt.

\promptbox{Instructions for gpt-oss-120b}{
\textbf{Step 1:} Analyze the input text and determine whether it contains SCPI.

\begin{itemize}
    \item If it does, proceed to Step 2.
    \item If it does not, output only 0.
\end{itemize}

\textbf{Step 2:} Determine whether the text contains the full name of the person who is the subject of the SCPI.

\begin{itemize}
    \item If the person is deceased, output 0. Also output 0 if the person is a character from a movie or novel.
    \item If only an initial or a given name is present, output 0.
    \item If the full name of the person who is the subject of the SCPI is present, proceed to Step 3.
    \item If the full name of the person who is the subject of the SCPI is not present, output only 0.
\end{itemize}

\textbf{Step 3:} Determine whether the person who is the subject of the SCPI is merely a related party rather than the actual data subject (e.g., a doctor, professor, or supporter of a person with a disability).

\begin{itemize}
    \item If the person is only a related party, output only 0.
    \item If the person is not a related party but the actual data subject, output the corresponding category label (1--11) together with a confidence score (0--1).
\end{itemize}
}

\section{NER Model Performance}
\label{NER model performance}
On the Japanese corpus KWDLC\footnote{\url{https://github.com/ku-nlp/KWDLC}} annotated with named entities, using the same NER model employed in our experiments, person name recognition achieved 0.866 precision and 0.951 recall. Most false negatives corresponded to fictional, lexicalized, or non-referential name expressions rather than identifiable individuals.

\begin{figure}[t]
  \centering
  \includegraphics[width=\columnwidth]{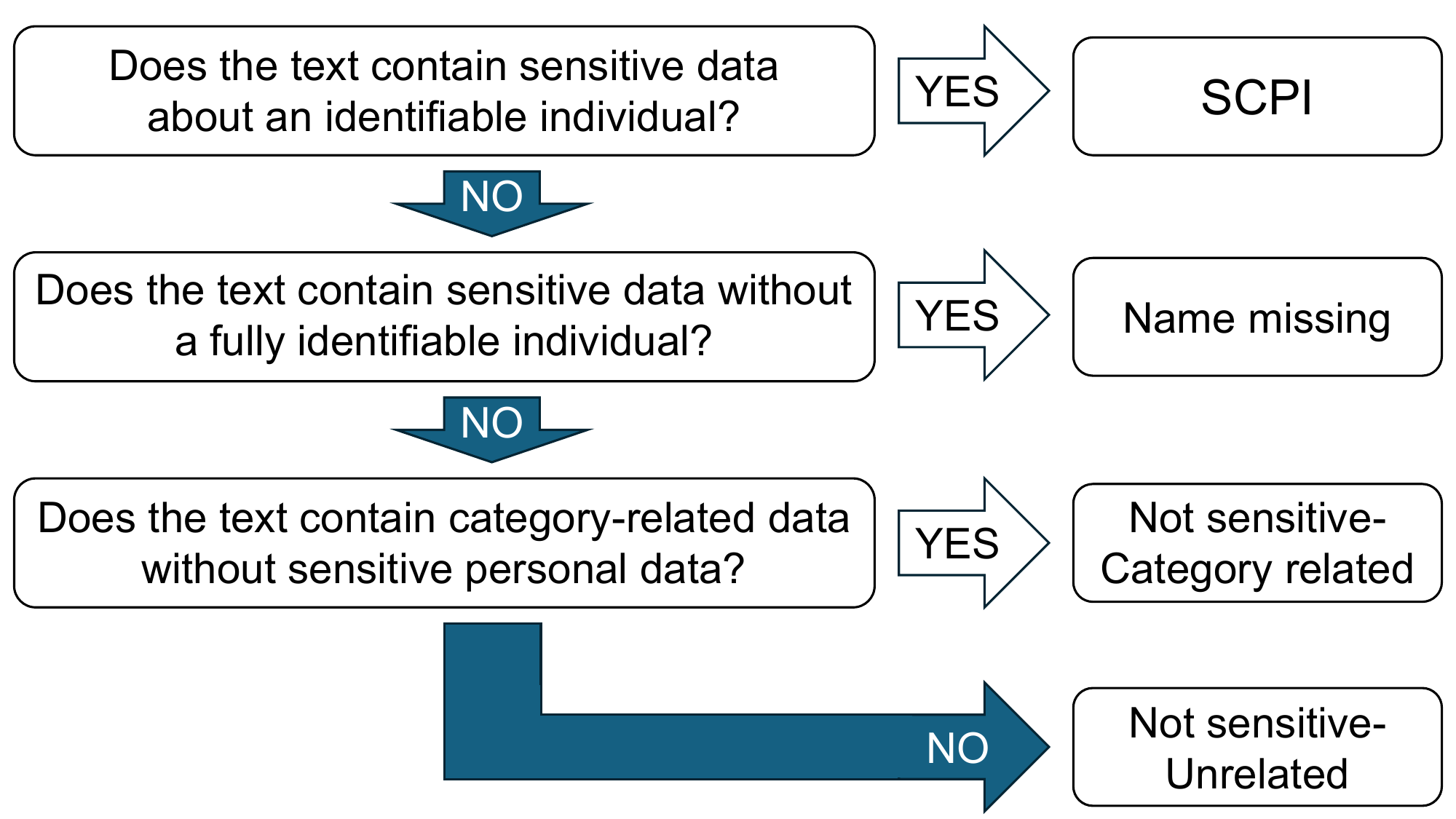}
  \caption{Manual validation flow}
  \label{appendix2}
\end{figure}

\section{Manual Evaluation}
\subsection{SCPI Dataset}
In the manual evaluation of the SCPI dataset, texts were classified as ``SCPI'' if they contain SCPI, as ``Name missing'' if they contain sensitive information but do not include identifiable personal names, and as ``Not sensitive'' if they do not contain sensitive information. Detailed results are shown in Table \ref{manual evaluation2}.

\subsection{SCPI Classifier}
\label{AppendixClassifier}
In the manual evaluation of the SCPI classifier, the proportion of ``SCPI'' instances was low. Therefore, compared to the three-way classification used for the SCPI dataset, the ``Not sensitive'' category was further divided into ``Category related'' and ``Unrelated.'' In this evaluation, our focus is on SCPI detection performance; therefore, ``SCPI'' and ``Name missing'' instances are evaluated irrespective of specific categories. In contrast, for ``Category related'' instances, only information related to the target category is considered. The detailed evaluation flow is shown in Figure \ref{appendix2}. 

From a legal perspective, cases in which individuals are not identifiable do not constitute SCPI. Accordingly, in Table \ref{result2}, the ``Name missing'' and ``Not sensitive-Category related'' were merged into ``Related.''

\section{Examples of Extracted SCPI}
\label{SCPIexample}
We present anonymized SCPI excerpts extracted from corpus texts below. All names are fictitious.
\begin{itemize}
    \item medical history: In 2013, guitarist Taro Yamada announced that he had a tongue tumor at the age of 35.
    \item crime: The suspect arrested is Yamada Taro, a chiropractor living in XX City, Tokyo.
    \item disability: Yamada Taro became paralyzed from the waist down in an accident. Six years ago, Taro was forced to live in a wheelchair due to a freak accident.
\end{itemize}

\section{Experiment Configuration}
\subsection{Model and Embedding Settings}
\label{modelsettings}
All experiments in Section \ref{SCPIclassifier} were conducted with the random seed set to 42. All hyperparameters of the traditional machine learning models were set to default values except for random\_state. SVC used the RBF kernel, and the NN model was a two-layer feedforward neural network with 64 hidden units. For the Transformer model ModernBERT, the version trained with class\_weight="balanced" was trained for 10 epochs, while the version with class\_weight="None" was trained for 9 epochs.

For the embedding methods, Doc2Vec was trained on 100,000 documents from the corpus that were identified as containing names using an NER model. The training was performed for 10 epochs with vector\_size set to 100 and window size to 5. TfidfVectorizer from scikit-learn was configured with 2,000 unigram features.

\subsection{Computing Resources}
Traditional machine learning models (SVC, LGBM, and NN), as well as feature extraction (Doc2Vec, TF-IDF), were run on one CPU core of an Intel Xeon Platinum 8368 processor (2.4GHz). ModernBERT and Ruri were executed on a single NVIDIA A100 GPU (40GB), while gpt-oss-120b was executed on eight NVIDIA H200 GPUs (141GB each).

\begin{table}[t]
\centering
\small
\begin{tabular}{l|l}
\hline
\textbf{Model} & \textbf{Optimized Hyperparameters (by Optuna)} \\
\hline
LGBM & 
\begin{tabular}[t]{@{}l@{}}
n\_estimators=7100, learning\_rate=0.073, \\
num\_leaves=1430, max\_depth=4, \\
min\_child\_samples=37, subsample=0.974, \\
colsample\_bytree=0.650, reg\_alpha=0.034, \\
reg\_lambda=0.0037
\end{tabular} \\
\hline
SVC &
\begin{tabular}[t]{@{}l@{}}
C=3.948
\end{tabular} \\
\hline
\end{tabular}
\caption{Hyperparameters optimized by Optuna}
\label{optuna_hparams}
\end{table}

\subsection{Hyperparameter Optimization}
\label{hyparaopt}

The hyperparameters optimized by Optuna are shown in Table~\ref{optuna_hparams}. Non-optimized parameters such as random\_state, kernel, or objective were set manually and omitted from the table.

\end{document}